\documentclass[10pt,twocolumn,letterpaper]{article}

\usepackage{iccv}
\usepackage{times}
\usepackage{epsfig}
\usepackage{graphicx}
\usepackage{amsmath}
\usepackage{amssymb}
\usepackage{subfigure}
\usepackage{rotating}
\usepackage{multirow}

\usepackage[pagebackref=true,breaklinks=true,letterpaper=true,colorlinks,bookmarks=false]{hyperref}

\iccvfinalcopy 

\def\iccvPaperID{807} 

\ificcvfinal\fi
\begin{document}

\title{Learning Multi-Scale Representations for Material Classification}

\author{Wenbin Li,  Mario Fritz\\
Max Planck Institute for Informatics, Saarbruecken, Germany\\
{\tt\small \{wenbinli, mfritz\}@mpi-inf.mpg.de}
}

\maketitle

\begin{abstract}
    The recent progress in sparse coding and deep learning has made unsupervised feature learning methods a strong competitor to hand-crafted descriptors. In computer vision, success stories of learned features have been predominantly reported for object recognition tasks. In this paper, we investigate if and how feature learning can be used for material recognition.  We propose two strategies to incorporate scale information into the learning procedure resulting in a novel multi-scale coding procedure. Our results show that our learned features for material recognition outperform hand-crafted descriptors on the FMD and the KTH-TIPS2 material classification benchmarks.
\end{abstract}

\section{Introduction}
Perceiving and recognizing material is a fundamental aspect of visual perception. It enables humans to make predictions about the world and interact with ease. In contrast to texture recognition it requires generalization over large variations between material instances and discriminance between visually similar materials. An efficient material recognition solution will find a wide range of uses such as in context awareness and robot manipulation. Studies have shown that material recognition in real-world scenarios is far from solved \cite{caputo05}. More recently, the task has been pushed even further to less constraint settings. The Flickr Material Dataset (FMD)  \cite{liu10cvpr} collects photos from Flickr as samples for common material and demonstrate the difficulties of material recognition.  In particular, they incorporate a large number of different descriptors in a Bayesian framework and provides a initial result on the dataset, yet well established manually designed feature descriptor, like LBP \cite{LBP96Ojala} and its variants \cite{Ojala2002,qi2012pairwise} have still been shown to be one  of the most powerful methods of feature descriptors and able to achieve state-of-art performance on the material recognition task. It is non-trivial to come up with a good design of visual features and efforts are clearly needed to explore the question how we can automatically learn features for this challenging and relevant problem.

Recent success of feature learning techniques \cite{Goodfellow+etal12:s3c} raises the question if the well established hand-crafted features in material recognition can be replaced with automatically learned ones. It is known that multi-scale representations are key for competitive performance on this task \cite{caputo05,leung2001representing,Ojala2002}. However, current feature learning techniques do not include multi-scale representations. Therefore, we investigate the applicability of different feature learning techniques to the material recognition task as well as how to bring multi-scale information to the feature learning process.

\begin{figure}
\begin{center}
\subfigure[Multi-scale filters learned on the KTH-TIPS2 database.]{
\label{fig:figures/kth_pyramids_selected}\includegraphics[width=0.28\linewidth]{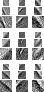}
}
\hspace{1cm}
\subfigure[Multi-scale filters learned on the FMD database.]{
\label{fig:figures/fmd_pyramids_selected}\includegraphics[width=0.28\linewidth]{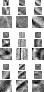}
}
\end{center}
\caption{
Example of the multi-scale filters learned on the KTH-TIPS2a (left) and the FMD (right) datasets. These represent a multi-scale code that is learned jointly based on the proposed MS4C model
}\label{fig:joint}
\end{figure}
        
\paragraph{Contributions}
We present the first study of applying unsupervised feature discovery algorithms for material recognition and show improved performance over hand-crafted feature descriptors. Further, we investigate different ways how to incorporate multi-scale information in the feature learning process. Hereby, we propose the first multi-scale coding procedure that results in a joint representation of multi-scale patches (see Figure \ref{fig:joint} for examples of multi-scale codes).
\section{Related Work}
\paragraph{Material Recognition}
Recognition of materials by appearance has received significant attention in the vision community. Curet database \cite{Dana1999} was first proposed to address the recognition problem of single material instance, which motivated a lot of progress on texture research \cite{texture05ijcv,varma2009statistical}. Later research \cite{Hayman04:OTS,caputo05} shifted the focus towards whole material class, emphasizing challenges like scale variation and intra-class variation. Liu et al \cite{liu10cvpr} presented the Flickr material dataset which used images from Flickr photos that were captured under unknown real-world conditions. Li et al \cite{Li2012} showed significant improvement over the previous results by only using a simple combination of color and Multi-scale LBP together with rendered data.  Hu et al \cite{hu2011toward} proposed the kernel descriptor \cite{bo2010kernel} and achieved state-of-art performance until recently Qi et al \cite{qi2012pairwise} proposed another variant of LBP descriptor to obtain further improvements over previous studies. All these efforts are based on hand designed descriptors while our approach investigates a learning based approach that starts from the raw pixel information.

\paragraph{Feature Learning}
One separate line of research is using learned features to tackle recognition problems. In typical supervised learning setting, one is given a set of examples $X=\{x_1,...,x_m\}$ and associated labels $\{y_1,...,y_m\}$. The goal is to learn a model to predict labels for new example $x$. The idea behind unsupervised feature discovery is to find a better representation $\phi(x)$ of data to ease the final learning problem. In the machine learning community, a rich set of models for feature discovery has been proposed. Examples includes sparse coding \cite{raina2007self}, restricted Boltzmann machines \cite{hinton2006fast,courville2011spike} and various autoencoder-based models \cite{bengio2007greedy}. The Spike-and-slab sparse coding (S3C) \cite{Goodfellow+etal12:s3c} has recently been proposed to combine the advantages of sparse coding with restricted Boltzmann machines and it has shown superior performance. We are based on the S3C model and show how to extend it to multi-scale feature learning as multi-scale feature representation is key in material recognition.

\paragraph{Multi-Scale Representation}
Already, the early texton work included multi-scale filters to enrich the representation. Although the clustering step can be seen as a form of feature learning, the filters are hand-crafted. Also the LBP work has seen extension to a multi-scale LBP \cite{Ojala2002} that has substantially improved the performance.  These feature extraction schemes are entirely hand-crafted.
Recently, a multi-scale convolutional neural network (CNN) \cite{farabet2012scene} trained from raw pixels to extract dense feature vectors that encode regions of multiple size centered on each pixel and then performed scene labeling has been proposed. It differs from our multi-scale feature learning approach, as we learn a representation jointly across scales. The image codes derived from our representation directly encode the multi-scaled information. Figure \ref{fig:joint} illustrates 12 of such multi-scale codes learned by our model.
\section{Feature Learning}
While we have seen broad application and success of feature learning techniques in object recognition, material recognition still relies on hand-crafted features. The appearance of material classes seem special in many ways. First of all, the samples seem to obey a stronger manifold assumption, as the appearance varies rather smoothly w.r.t. changes in lighting direction, orientation and scale. For objects, more drastic changes can occur due to the more pronounced 3D structure and edge information plays an important role.

In this section, we first described the framework for feature learning and then summarize several models we investigated for our tasks. Afterwards, we propose novel multi-scale feature learning strategies in order to accommodate for the multi-scale information that is important for material recognition.

\subsection{Unsupervised Feature Learning}
\begin{figure}
\begin{center}
\includegraphics[width=8.5cm]{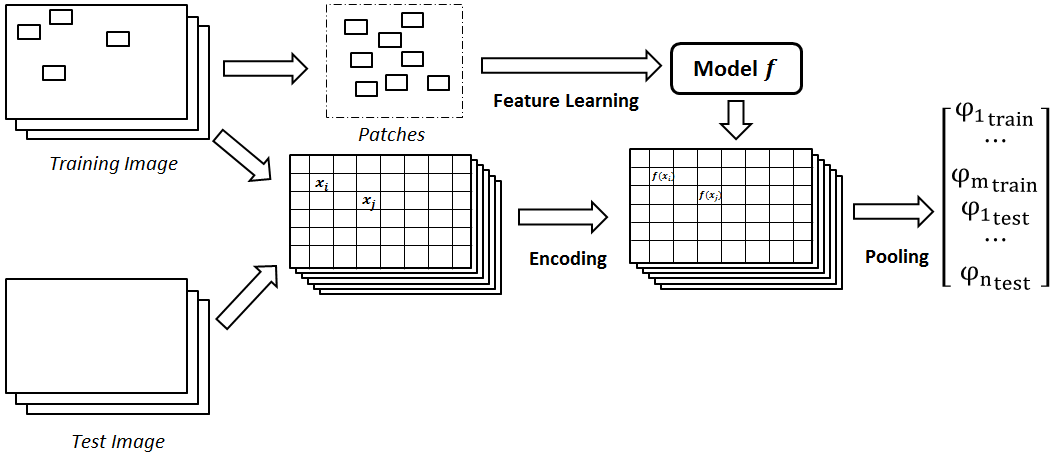}
\end{center}
\caption{Pipeline for unsupervised feature learning framework.}
\label{fig:pipeline}
\end{figure}

A commonly used patch-based unsupervised feature learning framework is illustrated in Figure~\ref{fig:pipeline}. First, random patches $\{v_1,...,v_n\}$ are extracted from training images and a feature mapping $f$ is learned (dictionary learning). Once the model is obtained, one can encode the patches covering the input image and pool the codes together in order to form the final feature representation (feature extraction). By altering the model used for feature mapping, we can get different feature representations.

\begin{figure}
\begin{center}
\includegraphics[width=1.8cm]{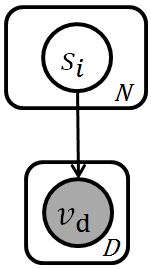}
\end{center}
\caption{Graphical model describing sparse coding (SC), where $v_d$ denote visible units, $s_i$ represent hidden variables and the directed arrows stand for dependency.}
\label{fig:sc}
\end{figure}

\paragraph{Sparse Coding (SC)}
Sparse coding for visual feature coding as illustrated in Figure~\ref{fig:sc} was originally proposed by Olshauen and Field \cite{olshausen1996emergence} as an unsupervised learning model of low-level sensory processing in humans. More recently, it was used in the self-taught learning framework \cite{raina2007self}. In the first phase, the dictionary $W$ -- also known as basis or codes -- is obtained by optimizing:

\begin{eqnarray*}
\begin{aligned}
& \underset{W,s_i}{\text{minimize}}
& & \sum_i \|v_i - W s_i\|_2^2 + \beta \|s_i\|_1 \\
& \text{subject to}
& & \|W_j\|_2 \leq 1, \forall j 
\end{aligned}
\end{eqnarray*}

Then in the next phase, feature representation $s_i$ for each input $v_i$ is obtained by solving the same form of optimization problem but with the learned dictionary.   

\begin{figure}
\begin{center}
\includegraphics[width=3.0cm]{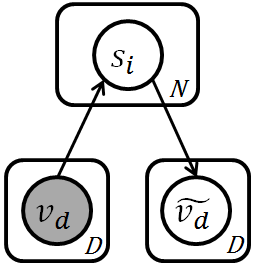}
\end{center}
\caption{Graphical model describing autoencoder (AE), where $v_d$ denote visible units, $s_i$ for hidden variables, $\widetilde{v_d}$ for reconstructed visible units and the directed arrows stand for dependency.}
\label{fig:ae}
\end{figure}

\paragraph{Auto-Encoder (AE)}
The Auto-encoder as illustrated in Figure~\ref{fig:ae} is another popular model widely used for learning feature representation in deep learning community. In the first phase, $v$ is  mapped into a latent representation $s$ (encoding) with a nonlinear function $f$ such as the sigmoid function:

\begin{eqnarray*}
s = f(W v + b)
\end{eqnarray*}

Then it is mapped back into a reconstruction $\widetilde{v}$ through a similar transformation: 
\begin{eqnarray*}
\widetilde{v} = f(\widetilde{W}s + \widetilde{b})
\end{eqnarray*}

and the dictionary (or weights) $W$ is obtained by optimizing the reconstruction error: 

\begin{eqnarray*}
W = \underset{W}{\operatorname{argmin}} \sum_i L(v_i,\widetilde{v_i})\\
\end{eqnarray*}

where $L(v_i,\widetilde{v})$ is a loss function such as the squared error $L(v,\widetilde{v}) = \|v-\widetilde{v}\|_2^2$.
Then during encoding phase, the features are computed by applying the forward-pass only in oder to obtain $s$.

\paragraph{ Spike-and-Slab Sparse Coding (S3C)}
The Spike-and-Slab Sparse Coding (S3C) by Goodfellow at el \cite{Goodfellow+etal12:s3c} has been recently proposed to combine the merits of feature learning methods like sparse coding and RBMs. 

The model is a two-layer generative process: the first layer is a  real-valued $D$-dimensional visible vector $v\in R^D$, where  $v_d$ corresponding to the pixel value at position d; the second layer consists of two different kinds of latent variables, the binary \emph{spike} variables $h\in\{0,1\}^N$ and the real-valued \emph{slab} variables $s\in R^N$. The spike variable $h_i$ gates the slab variable $s_i$, and those two jointly define the $i^{th}$ hidden unit as $h_i s_i$. The process can be more formally described as follows: 
\begin{eqnarray*}
\begin{aligned}
\forall i \in \{1,...,N\}&,d\in\{1,...,D\}\\
p(h_i=1)&=g(b_i)\\
p(s_i|h_i)&=N(s_i|h_i\mu_i,\alpha_{ii}^{-1})\\
p(v_d|s,h)&=N(v_d|W_{d:}(h \circ s),\beta_{dd}^{-1})\\
\end{aligned}
\end{eqnarray*}

\begin{figure*}[t]
\begin{center}
\subfigure[][S3C]
{
\includegraphics[width=0.20\linewidth]{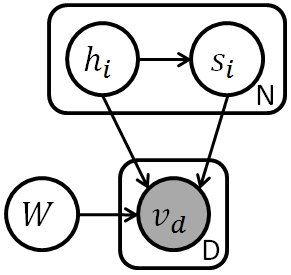}
}
\hspace{0.05\linewidth}
\subfigure[][S4C]
{
\includegraphics[width=0.22\linewidth]{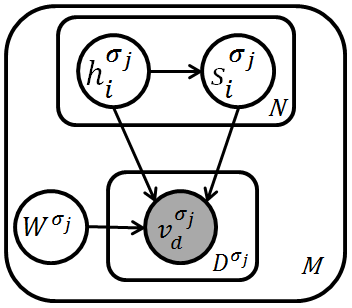}
}
\hspace{0.05\linewidth}
\subfigure[][MS4C]
{
\includegraphics[width=0.20\linewidth]{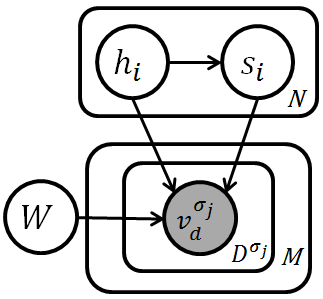}
}
\end{center}
\caption{Multi-scale feature models. (a) S3C, (b) S4C, (c) MS4C.}
\label{fig:multiscale}
\end{figure*}

where $g$ is the logistic sigmoid function, $b$ is a set of biases on the spike variables, $\mu$ and $W$ govern the linear dependence of $s$ on $h$ and $v$ on $s$ respectively, $\alpha$ and $\beta$ are diagonal precision matrices of their respective conditionals, and $h\circ s$ denotes the element-wise product of $h$ and $s$. Column of $W$ is constrained to have unit norm, $\alpha$ is restricted to be a diagonal matrix and $\beta$ to be a diagonal matrix or a scalar. In particular, W can be interpreted as a series of filters which can be used sparsely to represent the data. The graphical model describing it is shown in Figure~\ref{fig:multiscale} (a).

The  model has shown to outperform previous feature learning technique \cite{Goodfellow+etal12:s3c} and is the best performer on a recent transfer learning challenge \cite{transferworkshop} where one trains the model over the patches from the limited number of training images and a large number of unlabeled image data and then coded both training data and test data with the learned model, a standard linear SVM was then used for classification on the learned representation of data.

As discussed in \cite{Goodfellow+etal12:s3c}, one drawback of sparse coding is that the latent variables are not only encouraged to be sparse, but also to be close to 0 when activated. To tackle this issue, the S3C model introduces separate priors to control the activation of units and the magnitude of activated units separately. Though a similar structured RBM model known as $\mu-ssRBM$ \cite{courville2011unsupervised} is also proposed for feature learning, the non-factorial posterior of S3C model can grants better discriminative capability by selectively activating only a small set of features for a given input.   


\paragraph{Model Learning}
Variational EM \cite{saul1996exploiting} algorithm is used for model learning. It is a variant of EM algorithm with modification in the E-step where we only compute a variational approximation to the posterior rather than the posterior itself. In detail, the variational E-step maximize the energy functional with respect to a distribution $Q$ over the unobserved by minimize the Kullback-Leibler divergence: $D_{KL}(Q(h,s)||P(h,s|v)))$, where $Q(h,s)$ is drawn from a restricted family of distributions to ensure that $Q$ is tractable. For more details we refer the reader to \cite{Goodfellow+etal12:s3c}.

\section{Multi-Scale Feature Learning}
Scale information is a critical element for material and texture recognition problem. 
\cite{caputo05} showed that explicit treatment of scale is necessary for material recognition in realistic settings.
In \cite{Li2012}, Li et al performed a manifold alignment with respect to scale between real and synthesized data, which turned to be crucial for using the generated data to improve recognition rate. Similarly, local descriptors like LBP are limited by its small spatial support area, several extensions \cite{Ojala2002,maenpaa2003multi} for multi-scale descriptor have also been shown to yield strong performance improvements.  Therefore we propose two different strategies to include multi-scale information in feature learning:



\subsection{Stacked Spike-and-Slab Sparse Coding (S4C)}
In the first strategy, we perform the encoding at multiple scales and stack the obtained codes, then use this code for classification. We convolve the patch with different sized Gaussians before encoding in order to represent scale information. While there is a common dictionary, the representation already encodes how the patch evolves in scale-space and therefore multi-scale information is captured.
The graphical model describing it is shown in Figure~\ref{fig:multiscale} (b):
\begin{eqnarray*}
\begin{aligned}
\forall i \in \{1,...,N\}&,j\in\{1,...,M\}, d\in\{1,...,D\}\\
p(h_i^{\sigma_j}=1)&=g(b_i^{\sigma_j})\\
p(s_i^{\sigma_j}|h_i^{\sigma_j})&=N(s_i^{\sigma_j}|hi^{\sigma_j}\mu_i^{\sigma_j},(\alpha_{ii}^{\sigma_j})^{-1})\\
p(v_d^{\sigma_j}|s^{\sigma_j},h^{\sigma_j})&=N(v_d^{\sigma_j}|W_{d:}^{\sigma_j}(h^{\sigma_j} \circ s^{\sigma_j}),(\beta_{dd}^{\sigma_j})^{-1})\\
\end{aligned}
\end{eqnarray*}
where $M$ denotes the number of scales and $\sigma_j$ indexes units and parameters at specific scale.

\subsection{Multi-Scale Spike-and-Slab Sparse Coding (MS4C)}
In the second strategy, we first construct a multi-scale pyramid for each image, apply the feature learning directly on the pyramid and then use the obtained codes for classification. In contrast to the S4C approach, the MS4C approach yields filters/codes that model each patch jointly across scales. . 
The graphical model describing it is shown in Figure~\ref{fig:multiscale} (c):
\begin{eqnarray*}
\begin{aligned}
\forall i \in \{1,...,N\}&,j\in\{1,...,M\},d\in\{1,...,D\}\\
p(h_i=1)&=g(b_i)\\
p(s_i|h_i)&=N(s_i|hi\mu_i,\alpha_{ii}^{-1})\\
p(v_d^{\sigma_j}|s,h)&=N(v_d^{\sigma_j}|W_{d:}(h \circ s),\beta_{dd}^{-1})\\
\end{aligned}
\end{eqnarray*}
where $v_d^{\sigma_j}$ denotes the joint representation of visible units at specific scale $\sigma_j$. Inference is carried out as in the S3C model as the different scales can be seen as a decomposition of a larger multi-scale patch that includes all the scales.
Figure~\ref{fig:joint} shows 12 filters that we have learned in this manner. Each filter reaches across 3 scales. 

\begin{figure*}
\begin{center}
\label{fig:figures/kthtips2a.png}\includegraphics[height=3.2cm]{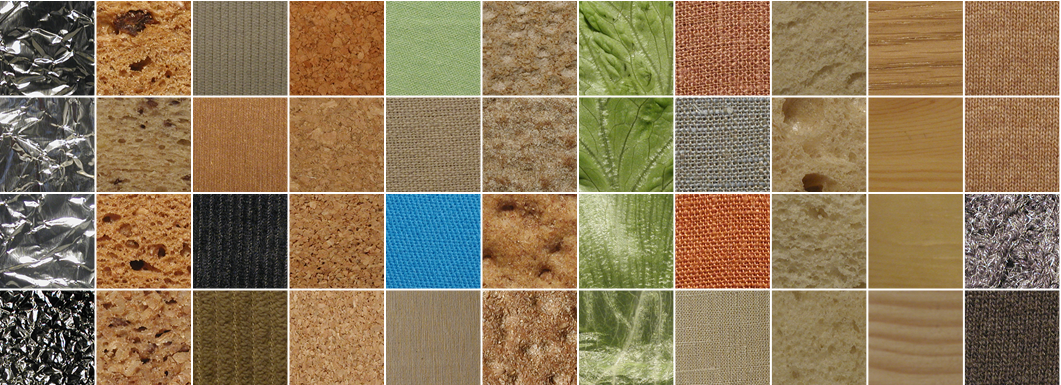}\hspace{0.3cm}\label{fig:figures/fmd.png}\includegraphics[height=3.2cm]{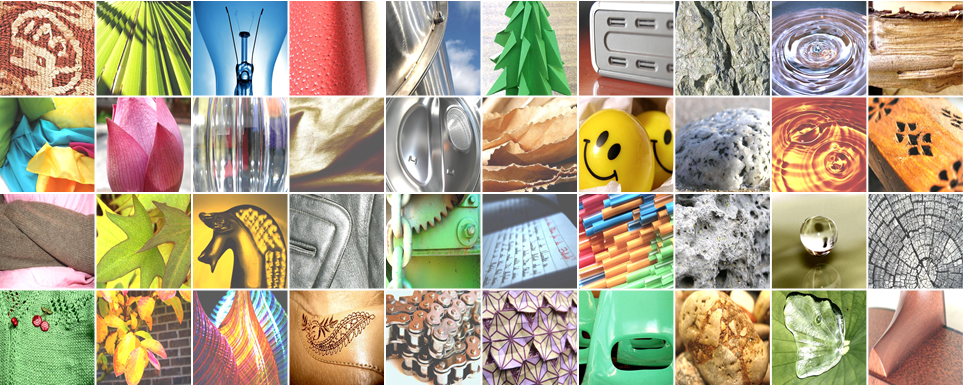}
\end{center}
\caption{The KTH-TIPS2a database (left) and the FMD database (right).}\label{fig:datasets}
\end{figure*}

\section{Experiments}
In our experiments, we investigate how the learning framework can be used for feature discovery on material recognition task and compare our approach to the  state-of-the-art on the FMD and the KTH-TIPS2 databases. Further we provide insights and visualizations on our learned representations.

\subsection{Datasets}
We use the KTH-TIPS2 database \cite{caputo05} and the Flickr Material Database (FMD) \cite{liu10cvpr} in our experiments. Example images are shown in Figure \ref{fig:datasets}. The KTH-TIPS2 database is designed to study material recognition with a special focus on generalization to novel instance of materials. It includes more than 4000 images from 11 material categories, and each category has 4 different instances. All the instances are imaged from varying viewing angles (frontal, rotated $22.5^\circ$ left and $22.5^\circ$), lighting conditions (from the front, from the side at $45^\circ$, from the top at $45^\circ$, and ambient light) and scales (9 scales equally spaced logarithmically over two octaves), which gives a total of $3\times4\times9=108$ images per instance. We use two instances for training and the other two for test per category. The FMD is collected from Flickr photos, including 10 common material categories with 100 images for each category, 1000 images in total. In our experiment, we randomly split half for training and the other half for testing as suggested in \cite{liu10cvpr}.


\subsection{Experimental setup }
We compare the learned features with hand-crafted features by the recognition rates on the two databases with standard SVM classifiers \cite{CC01a}. For single scale experiments, we compare to the LBP \cite{LBP96Ojala} and its several variants. For multi-scale approaches we consider: Texton \cite{leung2001representing}, Multi-scale LBP (MLBP) \cite{Ojala2002}. On the learning side, we compare to vector quantization, sparse coding, auto encoders and the spike-and-slab approach. In particular, we include a comparison with local quantization pattern (LQP)\cite{Hussain2012}-a recently introduced variant of LBP descriptor and kernel descriptor which has been shown the state-of-art performance on the FMD database. In all our experiments we fix the size of dictionary at 1600 for consistency.

\subsection{Single-scale}
For this group of experiments, we compare the performance between the learned features and the hand-crafted features.  In detail, for learned features, we apply the K-means clustering, Auto-Encoder (AE), Sparse Coding(SC) and the S3C model on the patch data where we vary the patch size of $\{6\times6, 12\times12, 24\times24\}$ (we had to skip the results for SC at patch size 24 as it turned out too costly in the encoding phase); for hand-crafted features, we examine the original LBP and several variants of LBP, including uniform-LBP ($LBP_u$), rotation invariant-LBP ($LBP_{ri}$) and rotation invariant, uniform-LBP ($LBP_{ri,u}$) as described in \cite{Ojala2002}. 

For implementation, we use Python and base on the library of Theano \cite{bergstra+al:2010-scipy} and Pylearn2 \cite{pylearn2} for the auto-encoder and the S3C model, which support GPU computation on network structure. For SC model, we use the SPAMS \cite{spams} package.

Experimental results are shown in Figure~\ref{tab:kth} and Figure~\ref{tab:fmd}. Each entry has the results for both the linear kernel (left) and the $exp-\chi^2$ kernel.
On both datasets, the S3C model in combination with the linear kernel outperforms all other hand-crafted and learned features. With a performance of $71.3\%$ and $48.4\%$ for the KTH-TIPS2a and the FMD respectively it improves by $4.1\%$ (over $LBP_u$ with the $exp-\chi^2$ kernel) and $9\%$ (over $LBP$ with the linear kernel) respectively. The best performance is achieved for a patch size of $12$. We verified that this parameter can be found via cross-validation on the training set. We attribute the decrease in the performance for the patch size of 24 to a lack of data to learn the required number of parameters. Best performance for feature learning technique is typically obtained in combination with linear kernel, while the hand-crafted features have to rely on the non-linear $exp-\chi^2$ kernel. This is another appealing property of the learned features from a computational point of view.


Based on these results, we found that the S3C feature did perform better than other learning approaches and the hand-crafted features for the single-scale setting, and hence we further developed the S3C model to multi-scale approaches in the following experiments.

\subsection{Multi-scale}
For this group of experiment, we introduce scale information with two different models, the Stacked S3C model (S4C) and the joint Multi-scale S3C model (MS4C), as described in Section 4. In particular, we also investigate the combination of color information for the MS4C model where we concatenate the MS4C codes with the S3C code at the base patch size. For hand-crafted features, we include the Multi-scale LBP (MLBP) and also the texton with the MR8 filter \cite{varma2002classifying}. Though the MR8 filter has been proposed for a long time, it still shows relative good performance on similar recognition tasks \cite{caputo05} and hereby it is also used as a baseline results in our experiments. Furthermore, as the filter banks are manually designed and also contain filters at multiple scales, we count it as a multi-scale hand-crafted feature although the textons are also learned via proper clustering algorithm such as K-means. Experimental results are shown in Figure~\ref{tab:kth} (a), (c) and Figure~\ref{tab:fmd} (a), (c).

MLBP shows better performance than textons in our experiments.
While the S4C model produces slightly worse performance than the MLBP on KTH-TIPS2, we see an improvement of $1.4\%$ for the MS4C. Further including color information improves the performance to $70.5\%$ which is an overall improvement of $3.8\%$ over the best hand-crafted descriptor.
From the numbers on the FMD database, we see that our S4C and MS4C beat the best hand-crafted feature (MLBP) by $7.2\%$ and $8\%$ respectively. On this database, inclusion of color information does not yield additional improvements.
The new joint multi-scale coding of of the MS4C consistently improves over the stacked approach of the S4C model.

\paragraph{Further Comparison to State-of-the-Art Descriptors} 
As not all papers follow the same experimental protocol, we reproduced two additional settings in order to provide more points of comparison to the state-of-the-art.
We follow the protocol in \cite{Hussain2012} and take 3 samples of each class for training and the fourth for testing on the KTH-TIPS2-a data, and then report averages over 4 random partitions via a simple 3-NN classifier, feature learned by single scale S3C at patch size of 12x12 achieved 70.2\%, which is significantly better than the reported results of 64.2\% for LQP.  Also we did additional experiments on the FMD database, following the settings in \cite{hu2011toward}, i.e. performing 5 trials and computing the average, and with multi-scale collaborated representation, we got average recognition rate of 48.3\% and standard deviation of 1.8\%, which is comparable to the best single kernel descriptor with 49\%.

\begin{figure}
\begin{center}
\subfigure[][Hand-crafted Feature.]
{
\footnotesize
\begin{tabular}{|c|c|c|c|c|c|}
\hline
\multicolumn{6}{|c|}{ClassificationRate(\%)}\\
\multicolumn{4}{|c|}{Single-Scale} &\multicolumn{2}{|c|}{Multi-Scale}\\
LBP &$LBP_u$ &$LBP_{ri}$ &$LBP_{ri,u}$ &Texton & MLBP\\
\hline
58.7/64.8 & 60.3/67.2 &55.0/53.6 & 50.9/51.4 &54.0/58.9 & 66.7/66.1\\
\hline
\end{tabular}
}

\subfigure[][Standard Feature Learning.]
{
\begin{tabular}{|cc|c|c|c|c|}
\hline
\multicolumn{2}{|c|}{\multirow{2}{*}{ } } & \multicolumn{4}{|c|}{ClassificationRate(\%)}\\
\multirow{3}{*}{\begin{sideways}PatchSize\end{sideways}}& & KM & AE & SC & S3C\\
\cline{3-6}
 &6&60.6/64.8&54.3/48.6&60.8/64.8&63.8/57.5\\
 &12&58.4/65.5&49.6/44.2&66.0/64.8&71.3/66.0\\
 &24&58.3/65.0&48.9/39.1&\** &55.9/60.8\\
\hline
\end{tabular}
}
\subfigure[][Multi-scale Feature Learning.]
{
\begin{tabular}{|c|c|c|}
\hline
\multicolumn{3}{|c|}{ClassificationRate(\%)}\\
S4C&MS4C&MS4C+Color\\
\hline
65.6/58.6&68.1/66.6&70.5/69.3\\
\hline
\end{tabular}
}
\end{center}
\caption{Experimental results on the KTH-TIPS2a database, where the first number in each entry is the recognition rate with the linear kernel, the other one is the number with the $exp-\chi^2$ kernel.}
\label{tab:kth}
\end{figure}

\begin{figure}
\begin{center}
\subfigure[][Hand-crafted Feature.]
{
\footnotesize
\begin{tabular}{|c|c|c|c|c|c|}
\hline
\multicolumn{6}{|c|}{ClassificationRate(\%)}\\
\multicolumn{4}{|c|}{Single-Scale} &\multicolumn{2}{|c|}{Multi-Scale}\\
LBP & $LBP_u$ & $LBP_{ri}$ & $LBP_{ri,u}$ &Texton &MLBP\\
\hline
39.4/36 &38.2/36.2&34.2/35.6&27.8/31.8&29.4/35.6&41.4/42.0\\
\hline
\end{tabular}
}
\subfigure[][Standard Feature Learning.]
{
\begin{tabular}{|cc|c|c|c|c|}
\hline
\multicolumn{2}{|c|}{\multirow{2}{*}{ } } & \multicolumn{4}{|c|}{ClassificationRate(\%)}\\
\multirow{4}{*}{\begin{sideways}PatchSize\end{sideways}}& & KM & AE & SC & S3C\\
\cline{3-6}
 &6&29.2/38.0&37.6/25.0&34.8/30.8&42.6/39.2\\
 &12&26.0/39.6&32.4/25.0&39.4/26.4&48.4/41.8\\
 &24&26.8/37.2&29.2/22.0&\** &40.8/44.0\\
\hline
\end{tabular}
}
\subfigure[][Multi-scale Feature Learning.]
{
\begin{tabular}{|c|c|c|}
\hline
\multicolumn{3}{|c|}{ClassificationRate(\%)}\\
S4C&MS4C&MS4C+Color\\
\hline
49.2/42.2&50.0/41.0&48.8/43.2\\
\hline
\end{tabular}
}
\end{center}
\caption{Experimental results on FMD database, where the first number in each entry is the recognition rate with the linear kernel, the other one is the number with the $exp-\chi^2$ kernel.}
\label{tab:fmd}
\end{figure}

\subsection{ Representation Transfer}
Additionally we investigate transferring representations across databases. In detail, we fixed the patch size at 12 and trained single scale S3C on KTH-TIPS2 database and then encoded the FMD data for classification and vice versa. Combined with the results in Figure~\ref{tab:kth},~\ref{tab:fmd} and Table~\ref{tab:Transfer Representation}, we can see that when encoding the image data in KTH-TIPS2 with the model learned on FMD, the performance degrades, yet still outperforms single scale LBP and color-patch; when representing the FMD data with the model learned on KTH, the performance even improves over any of the single scale descriptors. This indicates that the features learned through the S3C model on specific dataset are actually eligible to capture some common characteristics which generalize to different data within similar context.
\begin{table}
\begin{center}
\begin{tabular}{|c|c|c|}
\hline
Feature& \multicolumn{2}{|c|}{ClassificationRate(\%)}\\
\hline
 & Linear Kernel & $exp-\chi^2$ Kernel\\
\hline
\hline
\multicolumn{3}{|c|}{Code learned on FMD, and represent KTH-TIPS2}\\
\hline
S3C$(12\times12)$ &65.98 &61.11 \\
\hline
\hline
\multicolumn{3}{|c|}{Code learned on KTH-TIPS2, and represent FMD}\\
\hline
S3C$(12\times12)$ &44.4 &43.2 \\
\hline
\end{tabular}
\end{center}
\caption{Results for Transfer Representation.}
\label{tab:Transfer Representation}
\end{table}

\subsection{Discussion}
\paragraph{Visualization of Models}


Figure~\ref{fig:joint} shows visualization of our proposed Multi-Scale Spike-and-Slab Sparse Coding model. We see how each filter has a multi-scale response. We looked at a larger range of such filters, which reveals some more interesting properties. Some of these filters have a very similar structure across scales, while other do vary strongly. This observation and the strong performance numbers in our experiments let us conclude that a multi-scale code indeed captures additional information about how edge structures propagate through scales.

\paragraph{Effect of Patch Size}
Feature learning results for single scale descriptors are dependent on the patch size.  The size of patch determines the locality of the descriptor and therefore affects how the descriptor can be generalized to different instances, and from our experience, there seems not be any overall optimal patch size, which suggests we may need to try several candidates for a specific dataset and select the best one for use. In our experiments, we found that the patch size can be chosen based on cross-validation on the training set. Furthermore, our multi-scale approach S4C and MS4C resolve this problem by learning the representation across multi-scales. 


\paragraph{Scale Information}
Most of time, we see improvements when incorporating scale information, however on the KTH-TIPS2 database, we find that a descriptor learned at single scale performs the best. This may be related to the properties of the specific dataset and also the nature of our designed multi-scale descriptor. Both strategies for our multi-scale descriptors involve some redundancy between every scale that may degrade the classification performance, in return, this redundancy also encodes the scale information by itself that could improve performance, and the final performance will be affected by these two factors jointly. As for the KTH-TIPS2 database, material images were taken under strictly controlled conditions, in particular, only 9 different scales for all the instances, so the improvement via scale information is very limited in this case while the redundancy still affect the classification rate negatively. This in particular explains why the two multi-scale descriptors which already incorporate the information in model with patch size of 12 get worse results than the  single-scale descriptor. In contrast, for the case of the FMD database, images were collected from Flickr photos in arbitrary conditions, scale information become significantly more important and surpass the influence from the redundancy, which makes the multi-scale descriptor beat any of its components at single scale. In real world application, it seems closer to the latter situation, thus the multi-scale descriptor is preferable in this sense. 

 To further validate our analysis, we design additional experiments by making use of only subset of training data on the KTH-TIPS2 data in order to align the settings with the FMD database. So compared with the standard settings in the KTH-TIPS2 database where training data covering all different scales appeared in both training and test partition, we only use images taken under some scales for feature learning. Note this setting also resembles the situation in the FMD database, where data are imaged under unknown conditions such that training data cannot include the same scale information in test data. In this way, we would like to see if the multi-scale feature learning can provide extra power over the basic feature learning model. The experimental results are shown in Table \ref{tab:single-scale vs multi-scale}. As we can see from the table, when  observed very limited scales in training data like only one to three scales, the MS4C indeed outperforms the basic S3C single scale model.
 
\begin{table}
\begin{center}
\begin{tabular}{|cc|c|c|}
\hline
\multicolumn{2}{|c|}{\multirow{2}{*}{ } } & \multicolumn{2}{|c|}{ClassificationRate(\%)}\\
 & & S3C & MS4C\\
\hline
\multirow{4}{*}{\begin{sideways}Scale\end{sideways}} 
 &\{5\}&64.7/64.3 &65.2/58.4\\
 &\{3,7\}&65.8/66.4&67.7/65.9\\
 &\{3,5,7\}&65.1/66.7&70.7/67.7\\
 &\{1,3,5,7,9\} &67.3/66.4 &67.7/66.5 \\
&\{1,2,3,4,5,6,7,8,9\} &71.3/66.0 &68.1/66.6 \\
\hline
\end{tabular}
\end{center}
\caption{Recognition rates with different number scales available in the training dataset on KTH-TIPS2a database. The original database covers 9 different scales spanning two octaves which are indexed by No. 1 to 9 here.}
\label{tab:single-scale vs multi-scale}
\end{table}

\paragraph{Color vs. Gray-scale}
While color information serves as an important cue for visual recognition, it could also lead to confusion, so we should be careful to incorporate the color information. It is interesting to compare the results for the two multi-scale joint representation, with one in gray-scale and the other in color: on the KTH-TIPS2 database, the color information led to an improvement over the gray-scale representation while the gray-scale version achieved the best performance on the FMD database. It could be explained by the large variation of color information in the FMD data which causes the confusion, whereas the color cue is simpler and more informative for classification on KTH-TIPS2.

\section{Conclusions}
We have investigated different feature learning strategies for the task of material classification. Our results match and even surpass standard hand-crafted descriptors. Furthermore, we extended feature learning techniques to incorporate scale information. We propose the first coding procedure that learns and encodes features with a joint multi-scale representation. The comparison of our learned features with state-of-the-art descriptors shows improved performance on standard material recognition benchmarks.

{\small
\bibliographystyle{ieee}
\bibliography{ref}
}

\end{document}